\documentclass[review]{elsarticle}

\usepackage{amsmath}
\usepackage{amssymb}
\usepackage{algorithm}
\usepackage{comment}
\usepackage{natbib}
\usepackage{mathtools}
\usepackage{graphicx}
\usepackage{tabularx}
\usepackage{adjustbox}
\usepackage{subcaption}
\usepackage{algorithmic}
\usepackage{varwidth}
\usepackage{xcolor}
\usepackage{booktabs}
\usepackage{pgfplotstable}

\usepackage{lineno,hyperref}
\modulolinenumbers[5]
\definecolor{RevCol}{RGB}{255, 0, 0}
\journal{peer review}

\makeatletter
\def\ps@pprintTitle{%
	\let\@oddhead\@empty
	\let\@evenhead\@empty
	\def\@oddfoot{}%
	\let\@evenfoot\@oddfoot}
\makeatother

\begin{document}
	
	\begin{frontmatter}
		
		\title{NBMLSS: probabilistic forecasting of electricity prices via Neural Basis Models for Location Scale and Shape}
		
		\author[mymainaddress]{Alessandro Brusaferri\corref{mycorrespondingauthor}}
		\ead{alessandro.brusaferri@stiima.cnr.it}
		\author[mymainaddress]{Danial Ramin}
            \author[mymainaddress]{Andrea Ballarino}
		\cortext[mycorrespondingauthor]{Corresponding author}
		\address[mymainaddress]{CNR, Institute of Intelligent Industrial Technologies and Systems for Advanced Manufacturing, via A. Corti 12, Milan, Italy}

		\begin{abstract}
		  Forecasters using flexible neural networks (NN) in multi-horizon distributional regression setups often struggle to gain detailed insights into the underlying mechanisms that lead to the predicted feature-conditioned distribution parameters. In this work, we deploy a Neural Basis Model for Location, Scale and Shape, that blends the principled interpretability of GAMLSS with a computationally scalable shared basis decomposition, combined by linear projections supporting dedicated stepwise and parameter-wise feature shape functions aggregations.
          Experiments have been conducted on multiple market regions, achieving probabilistic forecasting performance comparable to that of distributional neural networks, while providing more insights into the model behavior through the learned nonlinear feature level maps to the distribution parameters across the prediction steps.
		\end{abstract}
		
	\end{frontmatter}
	
	
	\section{Introduction}
	Probabilistic forecasting of hourly electricity prices in day-ahead power markets (PEPF) is a complex problem with a significant impact. Accurate predictions and reliable uncertainty quantification are essential for a diverse array of participants, including utilities, retailers, aggregators, and large consumers \citep{MASHLAKOV2021116405}. These enable informed decision-making in high-stakes scenarios such as trading strategies, resource scheduling, and optimal commitment by factoring in potential fluctuations and associated risks \citep{WAGNER2022100246}.
Moreover, electricity prices are characterized by high volatility and rapid changes driven by intricate factors, including distributed power demand, generation costs, and weather conditions \citep{CIARRETA2022}. The growing adoption of renewable energy sources, vital for mitigating global emissions, introduces additional complexities to this landscape \citep{MADADKHANI2024107241}. Furthermore, fluctuations in gas prices critically affect power plant operations and overall electricity pricing. 

A wide range of methods has been proposed over the past decades to characterize the inherent uncertainty in probabilistic forecasts of electricity prices, including simple prediction intervals, discrete conditional quantiles, and full distributional estimation \citep{BRUSAFERRI20191158}. We refer interested readers to recent reviews for a detailed treatment \citep{NOWOTARSKI20181548},\citep{LAGO2021116983}.
Nowadays, an increasing research interest can be observed in neural network-based approaches, leveraging their flexible mapping capabilities in the conditioning space and exploiting the increasingly available computational power and tools \citep{LAGO2021116983}. In this context, Distributional neural networks (DDNN) parameterizing flexible forms, such as the Johnson's SU, have been recently shown to provide a valuable support in characterizing the complex patterns exhibited by the target price distributions, including, e.g., sensible heteroskedasticity, fat tails and skewness \citep{MARCJASZ2023106843}.
However, these enhanced capabilities come with a limited degree of interpretability. The inherent black-box nature of these models restricts the ability to understand how specific forecasts are generated, which may hinder their adoption in high-risk environments. Crucially, the way in which these models shape output density parameters based on input features throughout the prediction horizon remains opaque to users. 

Post-hoc model-agnostic techniques such as LIME surrogate models and Shapley values extensions have been proposed to provide a certain degree of explainability through local approximations and feature importance \citep{TSCHORA2022118752}. Still, faithful global explanations of the complex latent computations performed by the original models are difficult to achieve \citep{rudin19}.
Hence, a distinct line of research studies aims to enhance transparency in neural networks by acting directly on the architecture. A representative class of approaches in this field is the Neural Additive Model (NAM) introduced by \citep{10.5555/3540261.3540620}. Following the well-known Generalized Additive Models (GAM) framework, NAMs leverage a linear combination of flexible neural networks, each attending to single features, thus elucidating the actual relationships to the outputs. Recently, authors in \citep{pmlr-v238-frederik-thielmann24a} have extended NAMs from the original point prediction to a broader distributional regression setup, leading to Neural Additive Models for Location, Scale, and Shape (NAMLSS). By iterating on the general class of GAMLSS \citep{hirsch2024onlinedistributionalregression}, as NAMs do on GAMs, NAMLSSs aim to reveal the specific effects of input features on the shapes of complex response distributions beyond the mean, including variance, skewness, and tailedness. To this end, NAMLSS leverages a specific neural network for each input feature or even for each distribution parameter when configured to resemble the GAMLSS form. Despite their flexible and interpretable feature-level representation capabilities, such NAMLSS designs suffer from expansive computational scaling under increasing conditioning space and multi-horizon setups, making them impractical for the target PEPF applications at hand. To the best of our knowledge, studies exploring extensions to the NAMLSS concept for addressing challenging probabilistic time series forecasting applications such as PEPF are still lacking in the literature.

Following the aforementioned developments and moving from reported open issues, the major scope of the present work is to contribute to the investigation of the general Neural Additive Model framework within the PEPF field. In particular, we aim to assess whether more interpretable architectural forms can reach the state of the art distributional neural networks (DDNNs) under consistent parameterizations.
To this end, we extend the NAMLSS model by leveraging a basis decomposition of the feature shape functions \citep{10.5555/3600270.3600882}. Specifically, a unique neural network is exploited to learn a set of basis functions shared across all features in a multi-task setup, which are then combined by linear projections supporting dedicated stepwise and parameter-wise shape functions aggregations. The whole model is trained end-to-end through backpropagation, following the multi-horizon time series framework commonly adopted in DDNNs. We label the developed model NBMLSS hereafter, standing for Neural Basis Model for Location, Scale and Shape.
The experiments were conducted using open datasets \citep{ALIYON2024132877} from multiple day-ahead markets with heterogeneous characteristics. 

	\section{Methods}
	\label{Methods}
\begin{figure}[b!]
\begin{center}
\includegraphics[width=0.6\linewidth]{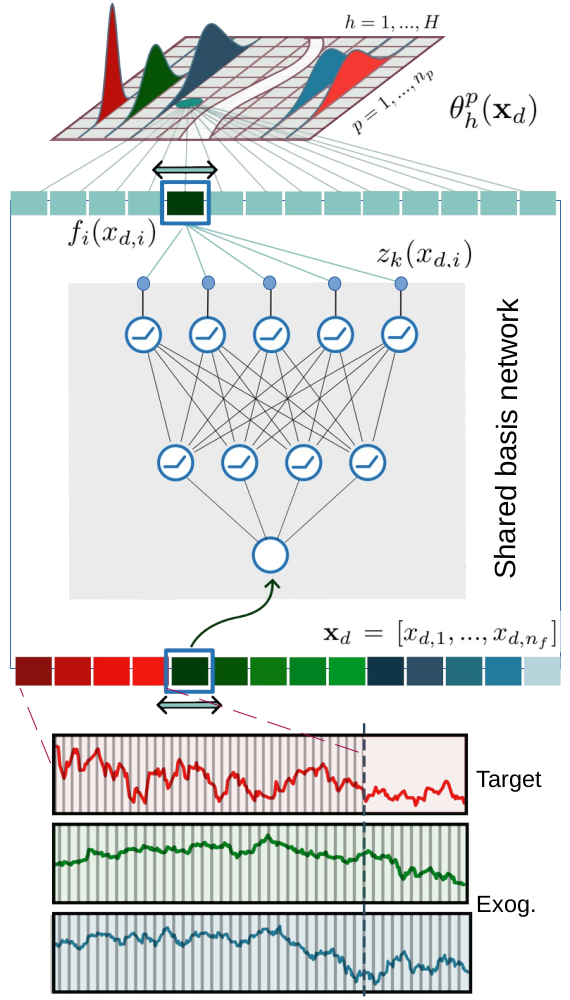}
\end{center}
\caption{NBMLSS schema}
\label{schema}
\end{figure}

As introduced in the previous section, this work is dedicated to the class of multi-step PEPF systems designed to learn the target price distributions over the different steps of the day-ahead horizon. $H$ represents the length of the target prediction horizon $\mathbf{y}_d=[y_{d,1},...,y_{d,H}]$ in each $d$-th day, e.g., next 24 hours. 
We concentrate on the feed-forward class (i.e., DNN) to align with the architecture of the original distributional neural network proposed in \cite{MARCJASZ2023106843} for PEPF.
The conditioning set is then structured as a flattened tensor $\mathbf{x}_d=[x_{d,1},...,x_{d,n_f}]$ of size $n_f$ typically involving the past values of the target price, the available dynamic covariates (e.g., generation and load forecasts) as well as constant features (e.g., day-of-week encoding, etc). 
We leave to future extensions the investigation of alternative architectural designs, such as hour-specific and recursive approaches.
Considering a dense feed-forward map with two hidden layers to shorten notation, the NBMLSS model architecture (depicted in Figure \ref{schema}) is formally defined as follows:
\begin{align}
z_{k}(x_{d,i})=& \mathbf{a}\left[ \sum_{j=1}^{n_u}\omega_{j,k}^{(2)} \mathbf{a} \left[\omega_{j}^{(1)}x_{d,i}\right]+\omega_{0,k}^{(2)}\right], k=1,...,n_z \label{bfe} \\
f_i(x_{d,i})=&\sum_{k=1}^{n_z}w_{i,k}z_{k}(x_{d,i}), i=1,...,n_f\\
\theta_h^p(\mathbf{x}_d)=&\mathbf{g}^p \left[\beta_h^p + \sum_{i=1}^{n_f}v^p_{h,i}f_i(x_{d,i}) \right], h=1,...,H; p=1,...,n_p
\end{align}

The first equation represents the $k$-th shared basis function \cite{10.5555/3600270.3600882} in the set of size $n_z \in \mathbb{Z}^+$. The second and third equations structure the stepwise target conditional distribution parameters $p=[1,...,n_p]$ as trainable linear projections from the basis output $z_{k}(x_{d,i})$ computed for each feature $x_{d,i}$. In particular, $w_{i,k}\in\mathbb{R}^{n_f \times n_z}$ is aimed to aggregate the shared basis into feature specific shape functions $f_i(.)$, $v^p_{h,i}\in \mathbb{R}^{H \times n_f}$ combines the shape functions outputs into the stage-wise distribution parameterization, and $\beta_h^p\in\mathbb{R}$ trainable biases. In the shared network Equation~\ref{bfe}, $\omega_{j}^{(1)}\in \mathbb{R}^{n_u}$ and $\omega_{j,k}^{(2)}\in \mathbb{R}^{n_u \times n_z}$ depict the trainable weights for the first and second hidden layer respectively, whereas $\mathbf{a}[.]$ represents the nonlinear activation functions. 
Although the vanilla NAM exploits exponential units activation to identify jagged maps \cite{10.5555/3540261.3540620}, subsequent studies have reported that standard ReLU provides a better alternative in practice \citep{bouchiat2024improvingneuraladditivemodels}.    
Dropout layers are also included on top of both hidden layers to foster basis decorrelation by random removal \cite{10.5555/3540261.3540620}.   
$\mathbf{g}^p[.]$ resembles the link function of the $p$-th parameter, designed depending on the adopted density form  \cite{pmlr-v238-frederik-thielmann24a}.
For instance, considering the Johnson’s SU density:
\begin{equation}
    d^h(\chi;\mathbf{x}_d)= \frac{\tau_d^h}{\sigma_d^h\sqrt{2\pi}}\frac{1}{\sqrt{1+\left(\frac{\chi-\lambda_d^h}{\sigma_d^h} \right)^2}}e^{-\frac{1}{2}\left[ \zeta_d^h + \tau_d^h \text{sinh}^{-1}\left(\frac{\chi - \lambda_d^h}{\sigma_d^h} \right) \right]^2}
\end{equation}
the link functions are computed as:
\begin{align}
\label{jsu_tr}
    &\mathbf{\Theta}(\mathbf{x}_d)=[\theta_1^1(\mathbf{x}_d),...,\theta_H^1(\mathbf{x}_d),...,\theta_1^{n_p}(\mathbf{x}_d),...,\theta_H^{n_p}(\mathbf{x}_d)]\\
    &\lambda_d^h = \mathbf{\Theta}(\mathbf{x}_d)^{[h]}\\
    &\sigma_d^h = \epsilon + \gamma\text{ Softplus}\left(\mathbf{\Theta}(\mathbf{x}_d)^{[H+h]}\right)\\
    &\tau_d^h = 1 + \gamma\text{ Softplus}\left(\mathbf{\Theta}(\mathbf{x}_d)^{[2\cdot H+h]}\right)\\
    &\zeta_d^h = \mathbf{\Theta}(\mathbf{x}_d)^{[3\cdot H+h]}\\
    &\text{Softplus}(x)=\text{log}\left(1+e^{x} \right)
\end{align}
where $\lambda_d^h,\sigma_d^h,\tau_d^h,\zeta_d^h$ are the density location, scale, tailweight and skewness conditioned on the input features $x_d$. $\epsilon=1e^{-3}$ and $\gamma=3$ are correction factors commonly introduced for computational purposes.
Similar transformations can be directly derived for further density parameterization, such as ensuring that standard deviation and degrees of freedom remain positive in the Student-t distributions.
The whole model is trained by backpropagation following a dynamic recalibration process, using the the negative log-likelihood computed on the output distribution as loss function. The samples $\mathcal{D}_n \equiv \{(x_d, y_d)\}_{d=1}^n$ are derived by applying standard moving window techniques on both conditioned/exogenous time series followed by concatenation. 
An early stopping procedure with patience is included by leveraging a validation subset. Additional details are provided in the next section, devoted to the experiment setup.

A further open issue in PEPF concerns the critical distribution shifts observed in recent power markets settlements. While models recalibration has been shown to provide partial mitigations, several research studies in the time series domain have reported reductions in point prediction errors by integrating Reversible Instance normalization (RevIN) layers \citep{kim2021reversible}. However, whether RevIN can provide valuable support to NN based PEPF systems beyond conventional dataset-level standardization techniques remains an open question requiring further investigation.
Consequently, we deploy RevIN within both NBMLSS and state of the art distributional NN architectures by introducing bijector layers aimed at reshaping the parameterized distributions back to the original scale.
	
	\section{Experiments and results}
	To evaluate the proposed approach, we leverage the open datasets recently structured by \citep{ALIYON2024132877}, which integrate data from the ENTSO-E transparency platform across several European markets. In particular, we focus on the day-ahead prices of Germany, Spain, Belgium, and Sweden-Stockholm (SE3), covering heterogeneous conditions. The investigation of additional datasets is reserved for future work. These datasets have been structured to include recent time periods (up to June 30, 2023) to support the assessment of model performance under the increasing volatility levels encountered by forecasters in contemporary contexts. In line with this objective, we have further extended them by incorporating records available on the transparency platform until September 30, 2024. 

All datasets include hourly load and wind/solar generation predictions as exogenous features. Besides, the temporal age feature and day-of-the-week indicator have been included, encoded in cyclical sine-cosine form.
As introduced above, the input/output samples are built from the time series by means of a standard moving window technique. For the scope of the present study, the conditioning vectors $x_d$ have been structured by integrating the price values over the previous days $d$-1, $d$-2, $d$-7 alongside the day-ahead exogenous variables, totaling $n_f$=147 features.

For each region, we have considered two out-of-sample testing conditions. The first, referred to as TS23, includes the last year of observations from the original datasets by \citep{ALIYON2024132877} (for consistency), spanning from July 1, 2022, to June 30, 2023. The second condition, labeled TS24, covers the subsequent year from October 1, 2023, to September 30, 2024.
In both cases, the initial training/validation subsets (before recalibration) consist of samples from January 1, 2019, up to the respective test dates. Following \citep{LAGO2021116983}, we used one year of data prior to each test set for hyperparameter tuning through cross-validation (via Optuna \citep{optuna}), with a random subsplit of 20\% adopted for early stopping. 

To balance the computational costs of updating the models under changing conditions across different experimental configurations, we employed weekly recalibration on the test set and retraining using four sequential folds during cross-validation.
Training is performed by means of Adam - minimizing the negative log likelihood loss (NLL) - with a maximum number of 800 epochs and a patience of 20 epochs. The batch size has been set to 128.

Considering the analysis performed in previous studies (see e.g., \citep{brusaferri2024onlineconformalizedneuralnetworks},\citep{MARCJASZ2023106843}), the Distributional Neural Networks (DDNNs) have been structured with two hidden layers, including ReLU activations and dropout regularization. Both Normal (N) and JSU (J) density parameterization are investigated. The number of units in each layer, the Dropout rate, as well as the learning rate have been tuned by grid search in the discrete sets [128, 512, 640, 768], [0, 0.1, 0.3, 0.5] and [1e-3, 5e-4, 1e-4, 5e-5] respectively. For NBMLSS, the number of units in the shared feature network has been searched within a reduced range $n_u,n_z$=[32, 64, 128, 256], as we observed that the procedure selected smaller values during the initial experiments on the datasets. Both models have been implemented using the TensorFlow Probability library \citep{tfp}.  
Table~\ref{hyper_tune_TS1} and Table~\ref{hyper_tune_TS2} (in Appendix) report the tuned configurations for each market under the testing conditions TS23 and TS24, respectively.    

Following \citep{MARCJASZ2023106843}, we have implemented both DDNN and NBMLSS within ensembles, aggregated either through uniform mixture (i.e., probability distribution aggregation - labeled '$p$') and quantile averaging (i.e., vincentization - labeled '$v$'). To this end, each model configuration has been recalibrated multiple times (i.e., 5), starting from different random initializations. The predicted distribution quantiles are estimated by generating 10000 samples for each test item. Post-hoc sorting is employed to fix potential quantile crossing. 

The probabilistic forecasting performances obtained on the test sets are reported in Tables~\ref{2023_table_BE}-\ref{2024_table_SE}, where the suffix '$r$' represents the backbone model including the RevIn layer, while '$s$' stands for the conventional Z-score normalization fitted on the training samples (see, e.g.,  \citep{brusaferri2024onlineconformalizedneuralnetworks})
The Continuous Ranked Probability Score (CRPS) has been approximated by the average Pinball loss across 99 percentiles.
We report the Kupiec test results for unconditional coverage (with significance level 0.05) on
50\% and 90\% prediction intervals beside the quantitative coverage probability (PICP), while the overall coverage achieved across the percentiles is depicted in the Figures~\ref{BE_cali}-\ref{SE_cali24}. Moreover, we show in Figure~\ref{DM_Pinball_fig},\ref{DM_Pinball_fig24}
the results of the multivariate Diebold and Mariano (DM) test on the differences in the CRPS loss norms between the models predictions. 

Firstly, it is noteworthy that the JSU parameterization has yielded superior performance compared to the Normal form in most cases, particularly for the more volatile TS23 test sets. This is in line with the observations reported in the previous studies on DDNNs, motivated by its greater flexibility in mapping complex aleatoric uncertainty patterns. Two exceptions to this ranking were observed in DE-TS24 and ES-TS24, which can be attributed to the more stable conditions in these cases.

Secondly, NBMLSS has achieved CRPS results that are comparable to, and in some cases better than, those of the DDNN under consistent settings (e.g., on BE with RevIN/standard normalization). 

\begin{table}[t!]
\caption{BE - TS23 test set results}
\label{2023_table_BE}
\begin{center}
\resizebox{\textwidth}{!}{
\begin{tabular}{lllllr}
&PICP$_{50\%}$(Kupiec) &PICP$_{90\%}$(Kupiec) &PICP$_{98\%}$(Kupiec) &MAE &CRPS 
\\ \hline \\
N-DNN$_{s,v}$    &58.0(6) &93.2(12) &97.7(22) &32.443 &11.819\\
N-DNN$_{s,p}$    &58.4(4) &94.1(8)  &98.2(22) &32.275 &11.805\\
N-DNN$_{r,v}$    &58.3(2) &91.2(20) &97.1(20) &26.184 &9.704\\
N-DNN$_{r,p}$    &58.7(2) &91.9(15) &97.6(20) &26.206 &9.709 \\
J-DNN$_{s,v}$    &50.3(20) &89.7(18) &96.9(16) &29.628 &10.714 \\
J-DNN$_{s,p}$    &51.6(18) &90.8(19) &97.5(21) &29.691 &10.726 \\
J-DNN$_{r,v}$    &51.4(20) &88.9(21) &96.6(17) &26.213 &9.535 \\
J-DNN$_{r,p}$    &52.2(19) &89.7(23) &97.1(20) &26.211 &9.534 \\
J-NBMLSS$_{s,v}$ &53.6(16) &90.3(21) &97.4(20) &28.242 &10.349 \\
J-NBMLSS$_{s,p}$ &53.7(16) &90.7(21) &97.7(21) &28.189 &10.338 \\
J-NBMLSS$_{r,v}$ &49.9(24) &87.8(19) &96.5(14) &25.350 &9.284 \\
J-NBMLSS$_{r,p}$ &50.1(24) &88.3(21) &96.8(18) &25.362 &9.282 \\
\end{tabular}
}
\end{center}
\end{table}
\begin{table}[t!]
\caption{DE - TS23 test set results}
\label{2023_table_DE}
\begin{center}
\resizebox{\textwidth}{!}{
\begin{tabular}{lllllr}
&PICP$_{50\%}$(Kupiec) &PICP$_{90\%}$(Kupiec) &PICP$_{98\%}$(Kupiec) &MAE &CRPS 
\\ \hline \\
N-DNN$_{s,v}$    &59.9(0)  &93.7(11) &97.8(17) &33.022 &11.968\\
N-DNN$_{s,p}$    &60.1(1)  &94.2(9)  &98.1(17) &32.969 &11.960\\
N-DNN$_{r,v}$    &54.2(12) &89.7(15) &96.2(14) &24.863 &9.094\\
N-DNN$_{r,p}$    &54.8(12) &90.5(15) &96.8(18) &24.860 &9.093 \\
J-DNN$_{s,v}$    &47.5(19) &87.2(14) &95.6(8) &25.872 &9.386 \\
J-DNN$_{s,p}$    &48.9(21) &88.7(17) &96.3(10) &25.940 &9.393 \\
J-DNN$_{r,v}$    &47.9(17) &87.1(15) &95.6(9) &25.350 &9.223 \\
J-DNN$_{r,p}$    &49.0(19) &88.1(14) &96.3(13) &25.347 &9.212 \\
J-NBMLSS$_{s,v}$ &53.2(18) &90.2(17) &97.0(17) &24.708 &9.019 \\
J-NBMLSS$_{s,p}$ &53.7(18) &90.8(18) &97.4(20) &24.649 &9.005 \\
J-NBMLSS$_{r,v}$ &46.5(17) &86.3(13) &96.2(12) &24.331 &8.834 \\
J-NBMLSS$_{r,p}$ &46.8(16) &86.9(14) &96.7(16) &24.333 &8.830 \\
\end{tabular}
}
\end{center}
\end{table}

\begin{table}[t!]
\caption{ES - TS23 test set results}
\label{2023_table_ES}
\begin{center}
\resizebox{\textwidth}{!}{
\begin{tabular}{llllll}
&PICP$_{50\%}$(Kupiec) &PICP$_{90\%}$(Kupiec) &PICP$_{98\%}$(Kupiec) &MAE &CRPS 
\\ \hline \\
N-DNN$_{s,v}$    &52.3(18) &87.6(17) &94.7(3) &16.788 &6.207\\
N-DNN$_{s,p}$    &53.3(17) &88.8(19) &95.6(8) &16.766 &6.196\\
N-DNN$_{r,v}$    &55.9(10) &89.4(21) &95.6(5) &16.211 &5.996\\
N-DNN$_{r,p}$    &56.3(9) &89.9(22) &96.2(13) &16.205 &5.995 \\
J-DNN$_{s,v}$    &47.1(19) &83.2(2) &92.9(0) &16.454 &6.139 \\
J-DNN$_{s,p}$    &48.4(21) &84.7(3) &93.8(0) &16.466 &6.126 \\
J-DNN$_{r,v}$    &49.1(20) &86.9(14) &95.3(5) &16.489 &6.044 \\
J-DNN$_{r,p}$    &49.8(20) &87.6(17) &95.9(7) &16.476 &6.038 \\
J-NBMLSS$_{s,v}$ &48.0(22) &83.4(3) &92.6(0) &16.675 &6.239 \\
J-NBMLSS$_{s,p}$ &48.3(23) &84.2(4) &93.6(0) &16.678 &6.233 \\
J-NBMLSS$_{r,v}$ &49.7(24) &87.9(16) &95.5(6) &16.308 &5.965 \\
J-NBMLSS$_{r,p}$ &50.2(24) &88.2(21) &96.0(8) &16.302 &5.962 \\
\end{tabular}
}
\end{center}
\end{table}
\begin{table}[t!]
\caption{SE - TS23 test set results}
\label{2023_table_SE}
\begin{center}
\resizebox{\textwidth}{!}{
\begin{tabular}{llllll}
&PICP$_{50\%}$(Kupiec) &PICP$_{90\%}$(Kupiec) &PICP$_{98\%}$(Kupiec) &MAE &CRPS 
\\ \hline \\
N-DNN$_{s,v}$    &54.7(14) &89.6(15) &94.8(7) &41.726 &15.794\\
N-DNN$_{s,p}$    &54.2(15) &90.5(18) &95.5(7) &41.469 &15.606\\
N-DNN$_{r,v}$    &56.9(9) &89.5(24) &96.1(12) &34.506 &12.813\\
N-DNN$_{r,p}$    &57.2(7) &90.5(22) &96.8(19) &34.377 &12.797 \\
J-DNN$_{s,v}$    &43.9(10) &83.1(5) &92.9(1) &39.040 &14.529 \\
J-DNN$_{s,p}$    &44.5(12) &84.2(5) &94.0(3) &39.262 &14.527 \\
J-DNN$_{r,v}$    &48.9(13) &85.3(7) &94.6(1) &33.498 &12.261 \\
J-DNN$_{r,p}$    &49.7(14) &86.1(9) &95.2(7) &33.494 &12.258 \\
J-NBMLSS$_{s,v}$ &47.9(22) &86.1(7) &94.3(0) &34.848 &13.063 \\
J-NBMLSS$_{s,p}$ &47.9(21) &86.5(9) &94.7(0) &34.423 &12.949 \\
J-NBMLSS$_{r,v}$ &45.7(14) &85.0(1) &94.6(0) &32.637 &11.999 \\
J-NBMLSS$_{r,p}$ &46.0(15) &85.5(3) &95.2(4) &32.606 &11.987 \\
\end{tabular}
}
\end{center}
\end{table}

\begin{table}[t!]
\caption{BE - TS24 test set results}
\label{2024_table_BE}
\begin{center}
\resizebox{\textwidth}{!}{
\begin{tabular}{lllllr}
&PICP$_{50\%}$(Kupiec) &PICP$_{90\%}$(Kupiec) &PICP$_{98\%}$(Kupiec) &MAE &CRPS 
\\ \hline \\
N-DNN$_{s,v}$    &49.7(24) &86.1(6) &94.4(2) &14.479 &5.309\\
N-DNN$_{s,p}$    &51.9(24) &88.6(23) &96.2(13) &14.489 &5.306\\
N-DNN$_{r,v}$    &56.0(6) &91.8(19) &97.4(22) &13.328 &4.860\\
N-DNN$_{r,p}$    &56.6(4) &92.8(14) &98.2(24) &13.342 &4.867\\
J-DNN$_{s,v}$    &45.4(15) &83.4(1) &93.4(0) &14.157 &5.188\\
J-DNN$_{s,p}$    &47.1(20) &85.3(4) &94.9(4) &14.145 &5.173\\
J-DNN$_{r,v}$    &48.5(20) &89.0(21) &97.0(19) &13.431 &4.847\\
J-DNN$_{r,p}$    &49.3(22) &89.9(22) &97.6(22) &13.426 &4.844\\
J-NBMLSS$_{s,v}$ &44.1(10) &84.7(3) &94.9(2) &13.632 &4.986\\
J-NBMLSS$_{s,p}$ &44.8(13) &85.4(4) &95.6(6) &13.615 &4.978\\
J-NBMLSS$_{r,v}$ &48.7(21) &87.9(19) &97.1(20) &12.758 &4.644\\
J-NBMLSS$_{r,p}$ &49.0(22) &88.3(21) &97.4(22) &12.758 &4.644\\
\end{tabular}
}
\end{center}
\end{table}
\begin{table}[t!]
\caption{DE - TS24 test set results}
\label{2024_table_DE}
\begin{center}
\resizebox{\textwidth}{!}{
\begin{tabular}{lllllr}
&PICP$_{50\%}$(Kupiec) &PICP$_{90\%}$(Kupiec) &PICP$_{98\%}$(Kupiec) &MAE &CRPS 
\\ \hline \\
N-DNN$_{s,v}$    &51.2(22) &89.2(18) &96.6(15) &10.300 &3.744\\
N-DNN$_{s,p}$    &54.1(17) &91.5(18) &97.7(22) &10.286 &3.746 \\
N-DNN$_{r,v}$    &60.9(1) &93.7(8) &98.1(24) &10.280 &3.785 \\
N-DNN$_{r,p}$    &61.4(1) &94.3(5) &98.5(23) &10.275 &3.789 \\
J-DNN$_{s,v}$    &52.0(18) &89.8(21) &97.0(15) &10.416 &3.778 \\
J-DNN$_{s,p}$    &54.1(14) &91.5(18) &97.6(22) &10.376 &3.776 \\
J-DNN$_{r,v}$    &54.3(13) &91.4(17) &97.8(23) &10.499 &3.809 \\
J-DNN$_{r,p}$    &55.9(8) &92.5(14) &98.4(24) &10.497 &3.814 \\
J-NBMLSS$_{s,v}$ &49.4(24) &87.9(17) &96.8(19) &10.729 &3.923 \\
J-NBMLSS$_{s,p}$ &50.8(23) &88.8(22) &97.3(22) &10.702 &3.917 \\
J-NBMLSS$_{r,v}$ &53.8(18) &91.3(24) &97.9(23) &10.230 &3.728 \\
J-NBMLSS$_{r,p}$ &54.2(17) &91.7(22) &98.1(24) &10.223 &3.727 \\
\end{tabular}
}
\end{center}
\end{table}
\begin{table}[t!]
\caption{ES - TS24 test set results}
\label{2024_table_ES}
\begin{center}
\resizebox{\textwidth}{!}{
\begin{tabular}{lllllr}
&PICP$_{50\%}$(Kupiec) &PICP$_{90\%}$(Kupiec) &PICP$_{98\%}$(Kupiec) &MAE &CRPS 
\\ \hline \\
N-DNN$_{s,v}$    &52.8(20) &91.0(23) &97.2(20) &11.362 &4.098\\
N-DNN$_{s,p}$    &54.5(15) &92.2(19) &97.7(24) &11.335 &4.099 \\
N-DNN$_{r,v}$    &57.9(3)  &91.1(21) &96.8(15) &11.607 &4.318 \\
N-DNN$_{r,p}$    &58.6(1)  &92.3(14) &97.5(18) &11.603 &4.325 \\
J-DNN$_{s,v}$    &46.2(15) &86.7(13) &95.7(8) &10.965 &3.955 \\
J-DNN$_{s,p}$    &48.7(21) &88.8(22) &96.7(15) &10.945 &3.947 \\
J-DNN$_{r,v}$    &53.2(17) &89.4(19) &96.2(13) &11.526 &4.253 \\
J-DNN$_{r,p}$    &54.0(14) &90.2(20) &96.7(16) &11.525 &4.253 \\
J-NBMLSS$_{s,v}$ &49.6(21) &88.4(24) &97.0(16) &11.276 &4.094 \\
J-NBMLSS$_{s,p}$ &50.0(23) &89.0(24) &97.4(22) &11.272 &4.093 \\
J-NBMLSS$_{r,v}$ &51.3(22) &88.6(20) &96.5(15) &11.242 &4.137 \\
J-NBMLSS$_{r,p}$ &51.6(23) &89.1(20) &96.7(16) &11.243 &4.136 \\
\end{tabular}
}
\end{center}
\end{table}

\begin{table}[t!]
\caption{SE - TS24 test set results}
\label{2024_table_SE}
\begin{center}
\resizebox{\textwidth}{!}{
\begin{tabular}{lllllr}
&PICP$_{50\%}$(Kupiec) &PICP$_{90\%}$(Kupiec) &PICP$_{98\%}$(Kupiec) &MAE &CRPS 
\\ \hline \\
N-DNN$_{s,v}$    &59.1(5) &94.3(5) &98.0(18) &12.196 &4.560\\
N-DNN$_{s,p}$    &60.0(4) &95.0(1) &98.3(17) &12.124 &4.553 \\
N-DNN$_{r,v}$    &63.8(0) &92.8(14) &97.2(17) &11.400 &4.351 \\
N-DNN$_{r,p}$    &64.1(0) &93.6(10) &97.7(22) &11.384 &4.347 \\
J-DNN$_{s,v}$    &52(22) &89.2(23) &96.2(13) &11.582 &4.275 \\
J-DNN$_{s,p}$    &53.5(16) &90.3(23) &97.1(17) &11.543 &4.268 \\
J-DNN$_{r,v}$    &53.3(15) &88.0(7) &95.6(9) &11.183 &4.151 \\
J-DNN$_{r,p}$    &53.9(15) &88.8(11) &96.1(9) &11.176 &4.149 \\
J-NBMLSS$_{s,v}$ &46.7(10) &87.1(12) &96.7(16) &11.673 &4.311 \\
J-NBMLSS$_{s,p}$ &49.0(11) &88.2(17) &97.3(20) &11.622 &4.296 \\
J-NBMLSS$_{r,v}$ &49.4(19) &88.1(17) &96.2(10) &10.861 &4.035 \\
J-NBMLSS$_{r,p}$ &50.2(20) &89.1(21) &96.8(17) &10.843 &4.029 \\
\end{tabular}
}
\end{center}
\end{table}

Across the different regions, the introduction of the RevIN layer appears to lead to lower CRPS values on average, beyond lower Mean Absolute Errors (on the extracted 0.5 quantile). 

Still, RevIN provides only partial mitigation, beyond the conventional dynamic recalibration, for strong drifts such as the one that affected most European power markets in 2022 following the Ukrainian conflict. Indeed, both N-DNN and J-DNN show excessively tails spikes in several stages during this period (see e.g., the negative spike around -400 euro/MWh in the upper part of Figure \ref{preds2022-2023}, reporting samples of the DE price in 2022). Similar problems with distributional neural networks in addressing sensible shifts have been reported in recent works \cite{LIPIECKI2024107934}. J-NBMLSS also appears to be critically affected, presumably because it is based on a restricted feature-wise mapping of the same network parameterization. The local normalization performed by RevIN is involved in such process, since shifting and centering observations sample-wise.  
Although the predicted distributions show better matches to the actual price settlements once the market dynamics return to more stable conditions (e.g., the lower part of Figure \ref{preds2022-2023}, reporting samples of the DE price in 2023), this issue warrants further investigation, e.g., by integrating alternative transformations, specific conditioning features, etc. 

\begin{figure}[t!]
\begin{center}
\includegraphics[width=1.0\linewidth]{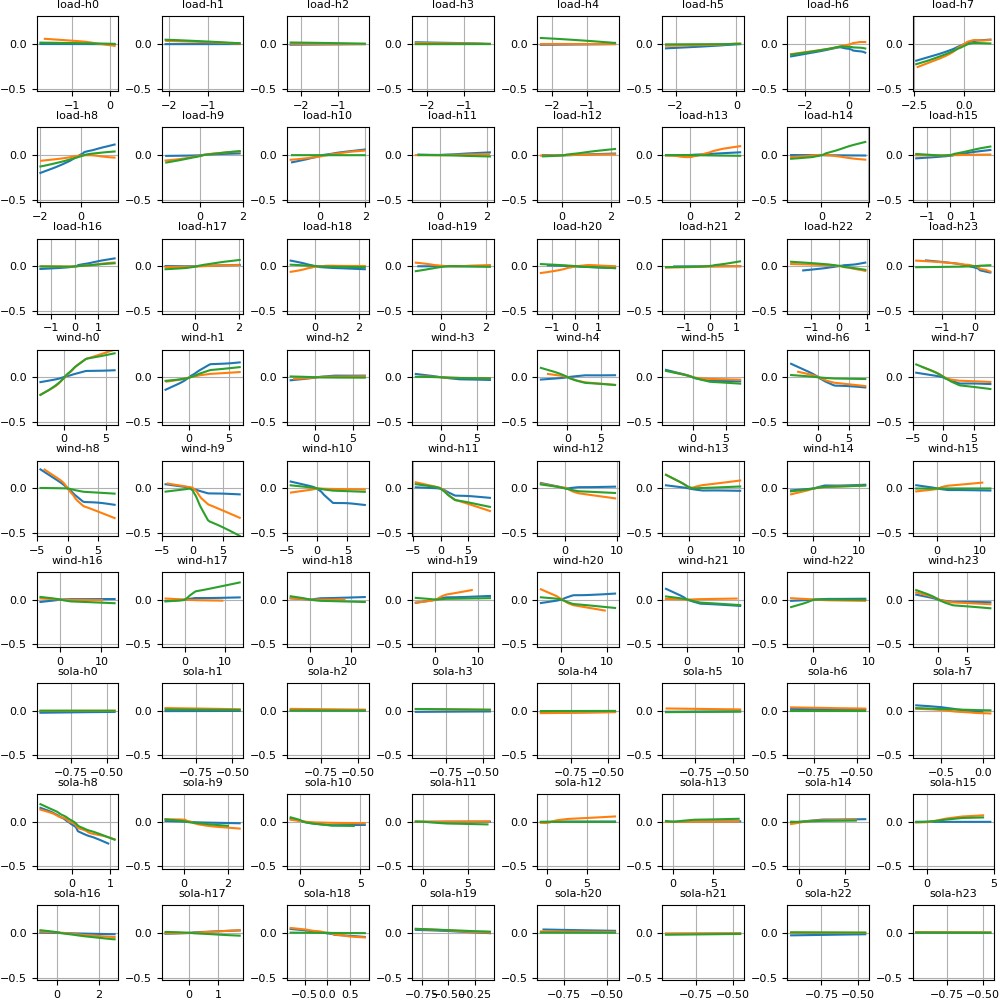}
\end{center}
\caption{Shape functions of 3 ensemble components on past price features (DE-h8-loc)}
\label{shape_futu_h8}
\end{figure}
\begin{figure}[t!]
\begin{center}
\includegraphics[width=1.0\linewidth]{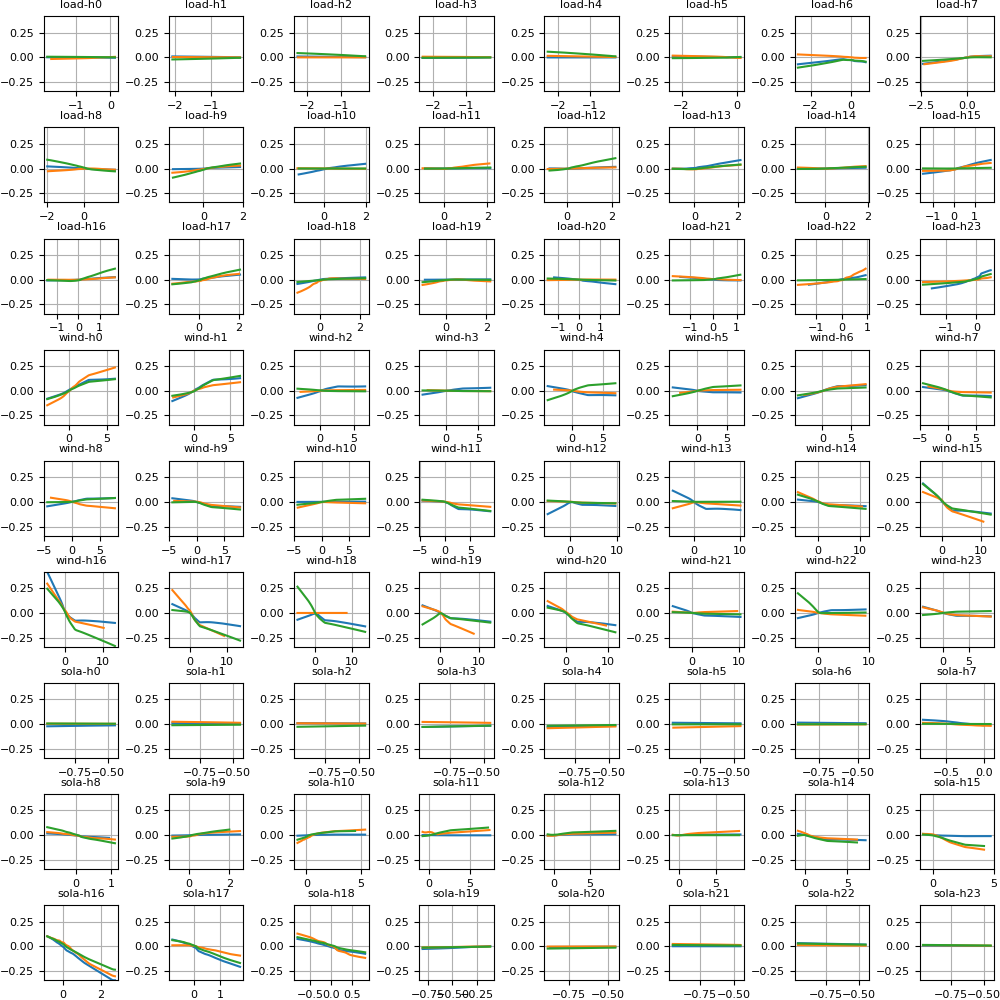}
\end{center}
\caption{Shape functions of 3 ensemble components on exogenous features (DE-h16-loc)}
\label{shape_futu_h16}
\end{figure}

\begin{figure}[t!]
\begin{center}
\includegraphics[width=1.0\linewidth]{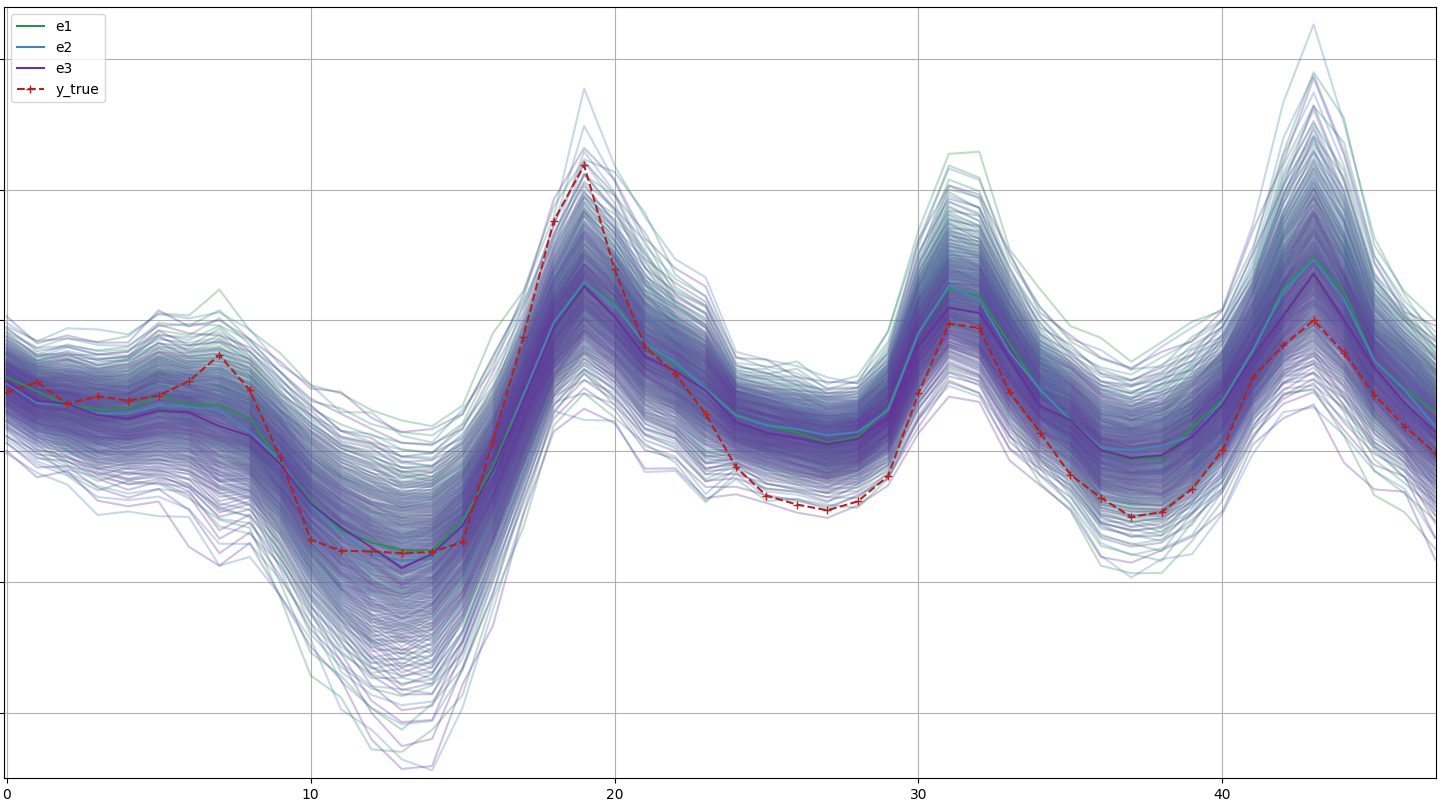}
\end{center}
\caption{Samples of predicted distribution deciles on DE market of 3 ensemble components}
\label{preds_ens}
\end{figure}

Moreover, we observe in Tables~\ref{2023_table_BE}-\ref{2024_table_SE} that both J-DNN and J-NBMLSS tend to produce partially overconfident probabilistic forecasts in multiple cases, in particular for the TS23 test sets, reflecting an open problem in neural network-based techniques \citep{Guo17}, \cite{10.5555/3666122.3667122}. 
The quantile averaging shows lower PICP values than the probability aggregation alternative, due to its typical ability to provide sharper forecasts \cite{Lichtendahl}. The specific impact on target coverage depends on the actual dispersion of the underlying ensemble components (see e.g., DE vs ES).
To address these issues, we plan to integrate ex-post conformal inference-based techniques in future extensions of this work.

Beyond the probabilistic forecasting performance achieved, the main aim of the NBMLSS architecture is to provide users with a representation of the relationships identified between each input feature and the target distribution parameters over the horizon, thereby revealing the underlying mechanisms driving to the output predictions.  
Figures \ref{shape_futu_h8}-\ref{shape_futu_h16} report samples of extracted shape functions from exogenous features to the location parameter of the JSU distribution for hours 8:00AM and 4:00PM in the prediction horizon, built by three random recalibration runs (i.e., ensemble components) over the last samples of TS24. 
Equal y-scales are employed to better display the relative contributions within the overall combination. 
Figure \ref{preds_ens} depicts the predicted percentiles from each component. 
It can be observed that the predicted distribution parameters are more influenced by the features corresponding to the same hour (see e.g., the relation to the wind, solar, and load inputs around hour 8.00AM versus 4.00PM in Figures \ref{shape_futu_h8} and Figures \ref{shape_futu_h16} respectively).  
Besides, it can be noticed that while the predicted distributions assume similar patterns in Figure \ref{preds_ens}, these are constructed by quite heterogeneous shape functions combinations (see e.g., the different shape functions for wind-h8 and wind-h9 in Figures \ref{shape_futu_h8}). 

Such latent behavior in neural additive models has also been reported in other recent studies showing the relation to the potential approximate concurvity among the features and the general under-specification issue of modern machine learning pipelines (see e.g., \cite{10.5555/3666122.3666956},\cite{Zhang24}, \cite{10.5555/3586589.3586815} and references therein).

\begin{figure}[t!]
\begin{center}
\includegraphics[width=1.0\linewidth]{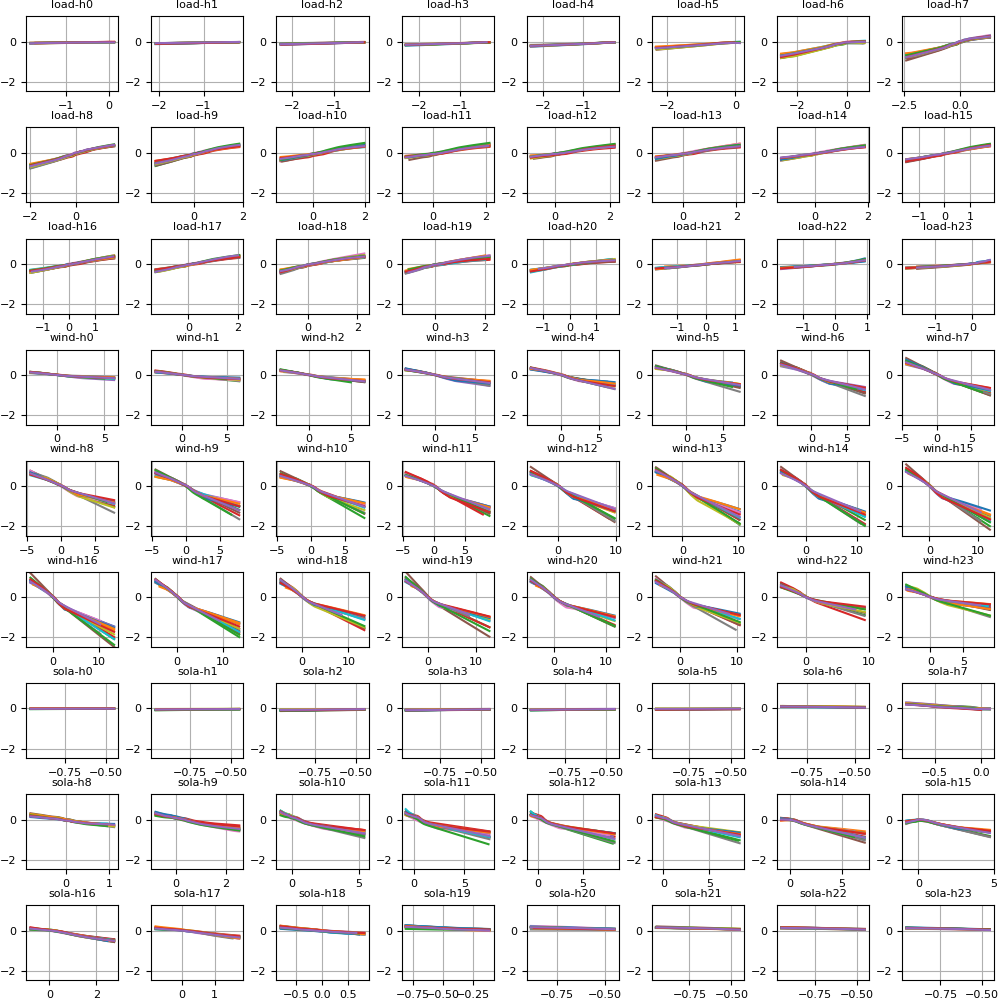}
\end{center}
\caption{Shape functions of the JSU location parameter on the masked exogenous subsets (3 ensemble components on DE)}
\label{shape_futu_sparse_loc}
\end{figure}

\begin{figure}[t!]
\begin{center}
\includegraphics[width=1.0\linewidth]{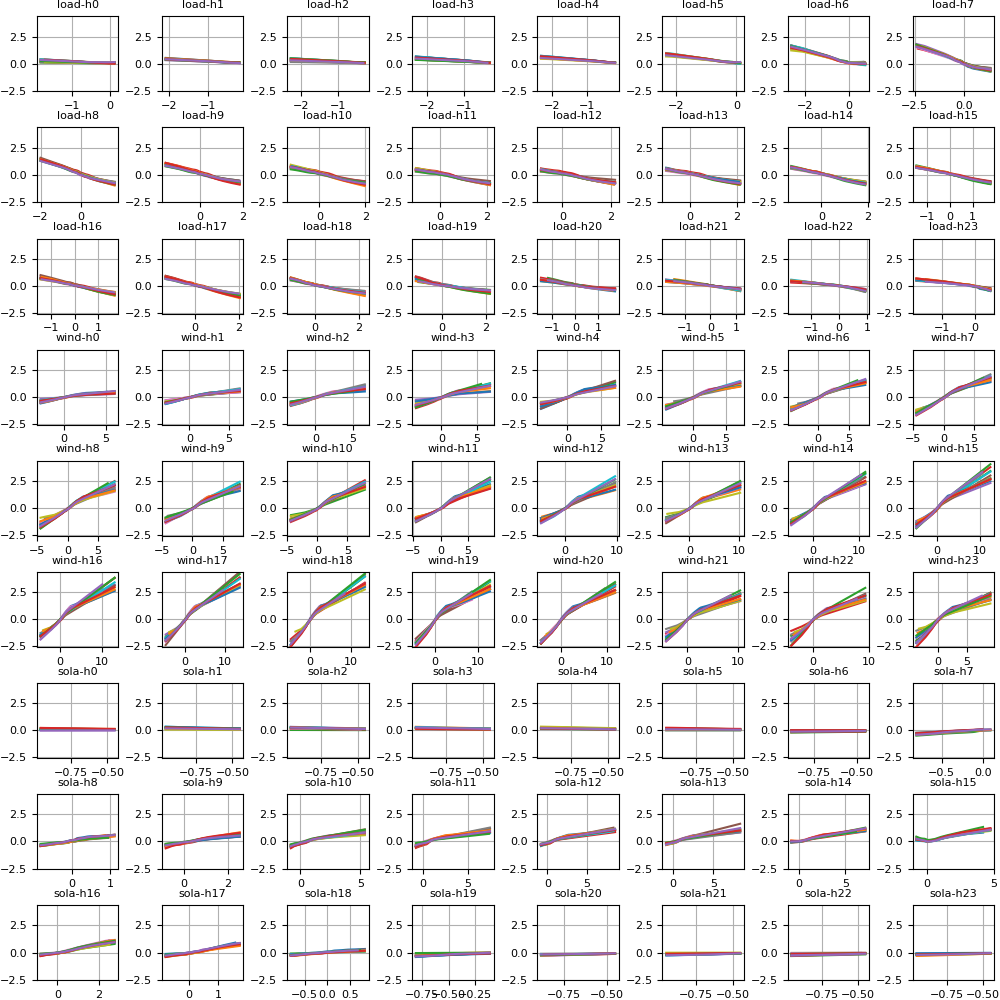}
\end{center}
\caption{Shape functions of the JSU skewness parameter on the masked exogenous subsets (3 ensemble components on DE)}
\label{shape_futu_sparse_skew}
\end{figure}

\begin{figure}[t!]
\begin{center}
\includegraphics[width=1.0\linewidth]{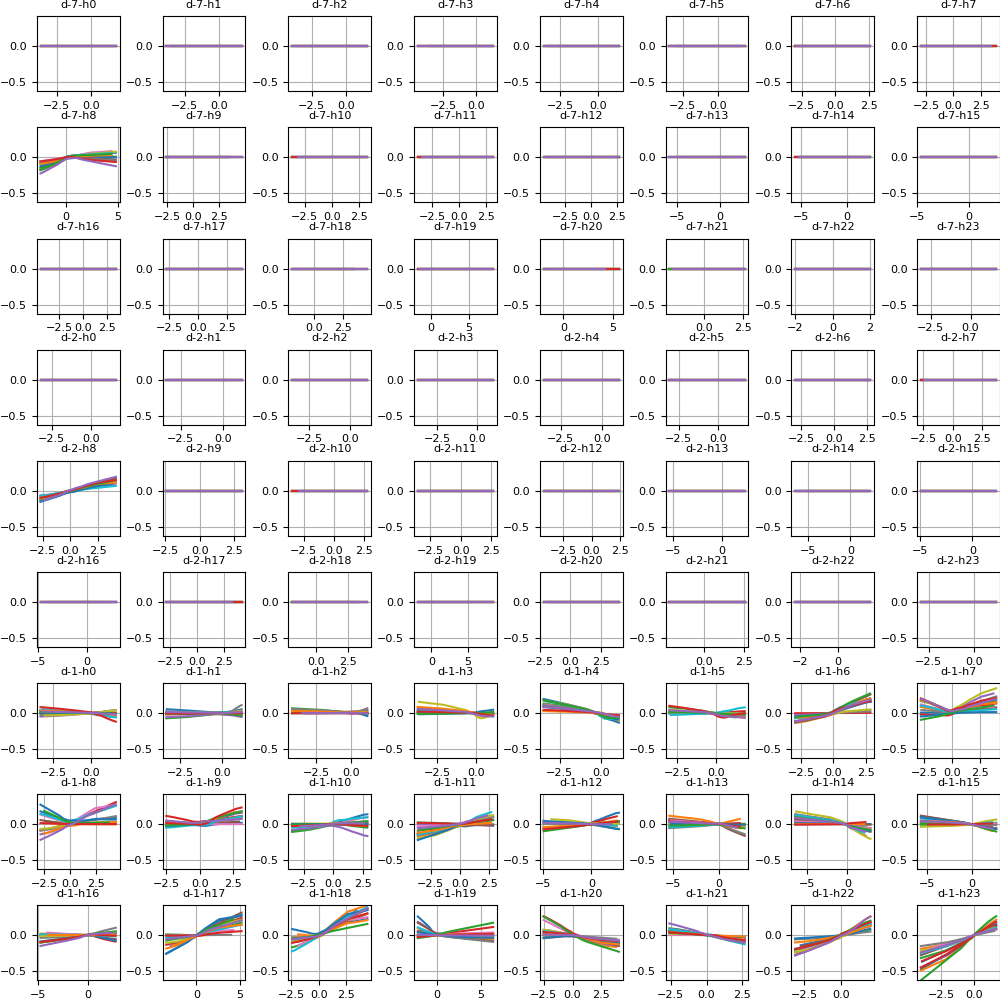}
\end{center}
\caption{Shape functions of the JSU location parameter at hour 8.00 to the past price values for the the masked exogenous configuration (3 ensemble components on DE)}
\label{shape_past_sparse_loc}
\end{figure}

To gain further insight, we conducted additional experiments with the JSU-NBMLSS architecture on TS24, utilizing a reduced number of features. In particular, based on the analysis performed in \cite{hirsch2024onlinedistributionalregression}, we masked the maps for each hour in the prediction horizon to exclude solar, wind, and load features from different hours. For instance, the JSU parameters at hour 12:00PM are conditioned solely on the solar, wind, and load forecasts from hour 12:00PM, while calendar and past price inputs are retained. 
Figure \ref{shape_futu_sparse_loc}, \ref{shape_futu_sparse_skew} report the shape functions, identified by 15 random trials, to the solar, wind and load features for both the location and skeweness \footnote{both easily displayable due to the identity link function applied} parameters across the prediction horizon, e.g., the subplot for load-h8 illustrates the shape function between the JSU parameter at 8:00 AM and the load at 8:00 AM. 
Besides, Figure \ref{shape_past_sparse_loc} displays the shape functions for the location parameter at hour 8:00 AM to the past price values.
The quantitative results obtained are reported in Table \ref{2024_table_red} and calibration plot \ref{TS24_cali_masked} with the label 'red'.

\begin{table}[t!]
\caption{TS24 test set results for the masked exogenous configurations}
\label{2024_table_red}
\begin{center}
\resizebox{\textwidth}{!}{
\begin{tabular}{llllllr}
& &PICP$_{50\%}$(Kupiec) &PICP$_{90\%}$(Kupiec) &PICP$_{98\%}$(Kupiec) &MAE &CRPS \\ \hline 
BE   &J-NBMLSS$_{r,v}^{red}$ &48.9(23) &89.0(22) &97.7(24) &13.245 &4.808\\
     &J-NBMLSS$_{r,p}^{red}$ &49.3(24) &89.5(22) &98(24) &13.242 &4.807\\
\hline 
DE   &J-NBMLSS$_{r,v}^{red}$ &54.1(19) &93.2(12) &98.5(23) &10.756 &3.916\\
     &J-NBMLSS$_{r,p}^{red}$ &54.4(17) &93.5(10) &98.6(23) &10.754 &3.917\\
\hline 
ES   &J-NBMLSS$_{r,v}^{red}$ &53.7(16) &90.7(16) &97.3(20) &12.053 &4.432\\
     &J-NBMLSS$_{r,p}^{red}$ &53.9(15) &91.0(17) &97.5(21) &12.049 &4.431\\
\hline 
SE   &J-NBMLSS$_{r,v}^{red}$ &48.1(17) &87.7(16) &96.9(19) &11.395 &4.259\\
     &J-NBMLSS$_{r,p}^{red}$ &48.7(15) &88.2(17) &97.2(19) &11.376 &4.254\\
\end{tabular}
}
\end{center}
\end{table}

First, we observe more stable patters across the runs as compared to the previous conditioning setup. Still, a sensible heterogeneity is present in the shape functions derived from the past price variables to the distributional parameters (see e.g., Figure \ref{shape_past_sparse_loc}), which are motivated as alternative representations with equivalent predictive capacity achieved by the learning pipeline \cite{10.5555/3666122.3666956}.   
Secondly, the load inputs show steeper relations to the JSU locations in the peak hours than during early morning and late evening, while having negative slopes on the skeweness. The renawable generation forecasts exhibit a typical shrinking influence on the price distribution locations and inverse relations to the skeweness parameters. Besides, sharper shapes are identified on the wind inputs within the central parts of the day, whereas the solar features assume expectable flat patterns during hours of minor irradiance. 
Finally, we observe deviations of the probabilistic forecasting performance in Table \ref{2024_table_red} due to the masked configuration, indicating a certain degree of exploitation of further features within the overall conditioning sets by the complete models.

This first analysis opens several avenues for further investigation in future extensions of the present study.
Rashomon sets approximation techniques \cite{10.5555/3666122.3668598} and automatic feature selection approaches controlling multivariate concurvity \cite{10.1007/s00180-022-01292-7} represent interesting directions to explore. Additionally, concurvity regularizers have been proposed to enforce decorrelation \cite{10.5555/3666122.3666956} within generalized additive architectures. However, nonlinear dependencies of shape functions have not yet been addressed in the literature, and the integration of additional penalties remains a compelling area for future research \cite{Zhang24}. Furthermore, deeper hyperparameter searches, such as through Bayesian optimization on expanded spaces, should be performed to gain further insights into the contribution of the various components within the entire learning pipeline to the final prediction performances. 

Overall, we concur with the authors in \cite{Zhang24} that optimal feature selection should be conducted under the supervision of domain experts, and that offering an ensemble of solutions, rather than a single candidate, currently appears to be the most practical trade-off.
This is particularly critical in time series applications, where redundancy in the conditioning set is more common than exceptional (as can occur, for example, among price values during early morning hours on closed days).
In this context, neural additive models such as NBMLSS can complement the flexibility of DDNN (e.g., through hybrid ensembles) by offering deeper insights into the underlying feature’s contribution across the domain, thereby supporting users during the critical phases of model design.

	\section{Conclusions and next developments}
	In this work, we have tackled probabilistic electricity price forecasting through Neural Basis Models for Location Scale and Shape, with the aim of assessing whether these more interpretable architectural forms can reach the state-of-the-art distributional neural networks (DDNNs). 
To enhance computational scalability under increasing conditioning space and multi-horizon setups, we have exploited a unique dense map aimed at learning a set of share basis combined by trainable linear projections into the stage-wise distribution parameterization.
Experiments have been conducted on open datasets from diverse power markets, incorporating recent time periods to support assessments under the increasing volatility levels that forecasters are facing. 
The proposed approach has achieved scores comparable, and in some cases improved, relative to those of the DDNN under consistent settings, while the Johnson’s SU obtained better performance than the conventional Normal form. This can be due to its greater flexibility in mapping complex aleatoric uncertainty patterns. Next, we have inspected samples of feature shape functions extracted by the model under different random initializations, pointing to the challenging concurvity problem. Still, we envision several avenues of future research, including the investigation of second-order interactions, the integration of eXplainable AI techniques (e.g., SHAP), further transformations as well as different components to better handle nonlinear dependencies within the conditioning sets. Moreover, we plan to investigate different probabilistic forecasting tasks, over both short and long term horizons, and network architectures.
		
	\bibliography{mybibfile_v01}

        \appendix
        
        \section{Appendix}
        \begin{table}[t!]
\caption{Hyperparameters selected by the grid-search procedure for TS23}
\label{hyper_tune_TS1}
\begin{center}
\small
\begin{tabular}{lllll}
\hline
\bf{N-DNN} &BE &DE &ES & SE 
\\ \hline
${n_u}$ &640  &768  &640  &512 \\ 
${l_r}$ &5e-4 &1e-4 &1e-4 &1e-3 \\
${d_r}$ &0.3  &0.3  &0.3  &0.1 \\ \hline
\bf{J-DNN} &BE &DE &ES & SE 
\\ \hline
${n_u}$ &640  &768  &768  &640 \\ 
${l_r}$ &5e-5 &1e-4 &5e-5 &5e-5 \\
${d_r}$ &0.1  &0.3  &0.3  &0.3 \\ \hline
\bf{J-NBMLSS} &BE &DE &ES & SE 
\\ \hline
${n_u}$ &64   &128  &32   &128 \\ 
${l_r}$ &5e-4 &5e-4 &5e-4 &5e-4 \\
${d_r}$ &0.5  &0.5  &0.3  &0.5 \\
\end{tabular}
\end{center}
\end{table}
\begin{table}[t!]
\caption{Hyperparameters selected by the grid-search procedure for TS24}
\label{hyper_tune_TS2}
\begin{center}
\small
\begin{tabular}{lllll}
\hline
\bf{N-DNN} &BE &DE &ES & SE 
\\ \hline
${n_u}$ &512  &768  &640  &768 \\ 
${l_r}$ &1e-3 &5e-5 &1e-3 &1e-3 \\
${d_r}$ &0.3  &0.3 &0.3  &0.3 \\ \hline
\bf{J-DNN} &BE &DE &ES & SE 
\\ \hline
${n_u}$ &768  &512  &640  &762 \\ 
${l_r}$ &5e-4 &5e-4 &1e-4 &1e-4 \\
${d_r}$ &0.3  &0.3  &0.3  &0.3 \\ \hline
\bf{J-NBMLSS} &BE &DE &ES & SE 
\\ \hline
${n_u}$ &128   &64  &32   &64 \\ 
${l_r}$ &5e-4 &1e-4 &5e-4 &5e-4 \\
${d_r}$ &0.5  &0.3  &0.3  &0.1 \\
\end{tabular}
\end{center}
\end{table}

\begin{figure}[t!]
\begin{center}
\includegraphics[width=0.91\linewidth]{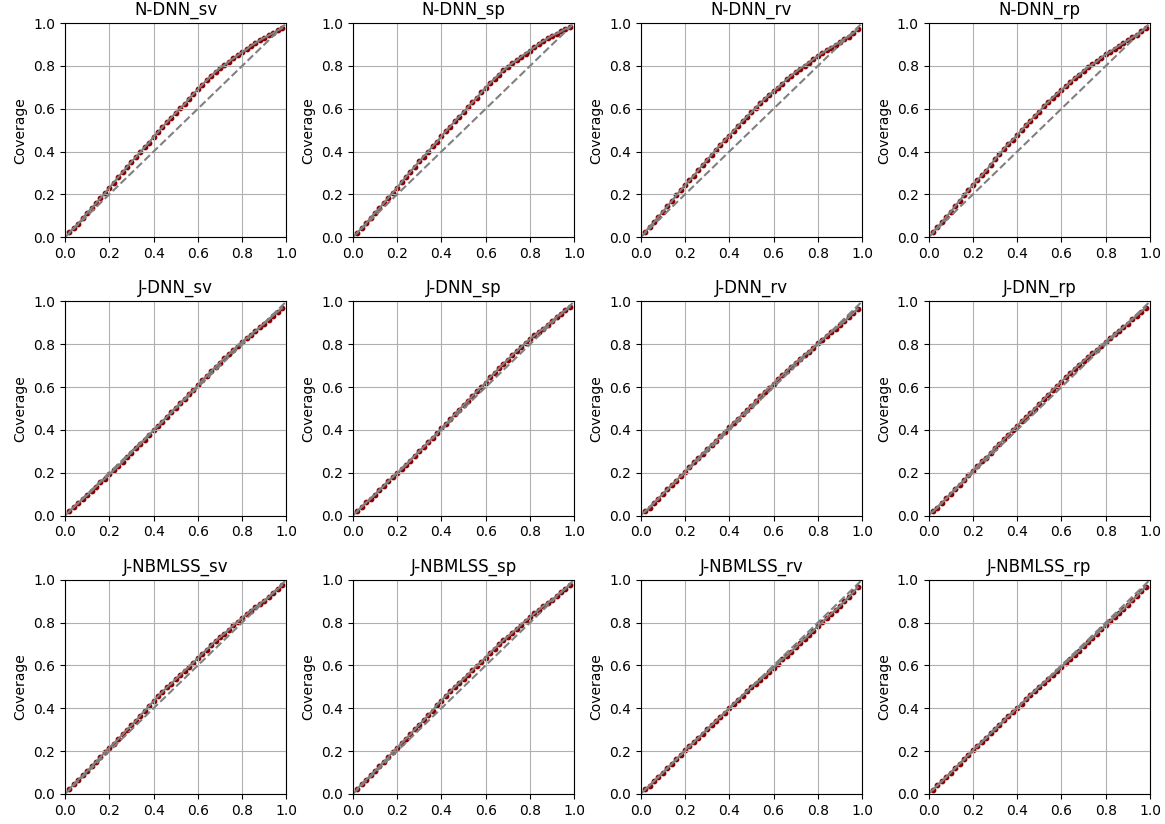}
\end{center}
\caption{Calibration plots - BE TS23 test set}
\label{BE_cali}
\end{figure}
\begin{figure}[t!]
\begin{center}
\includegraphics[width=0.91\linewidth]{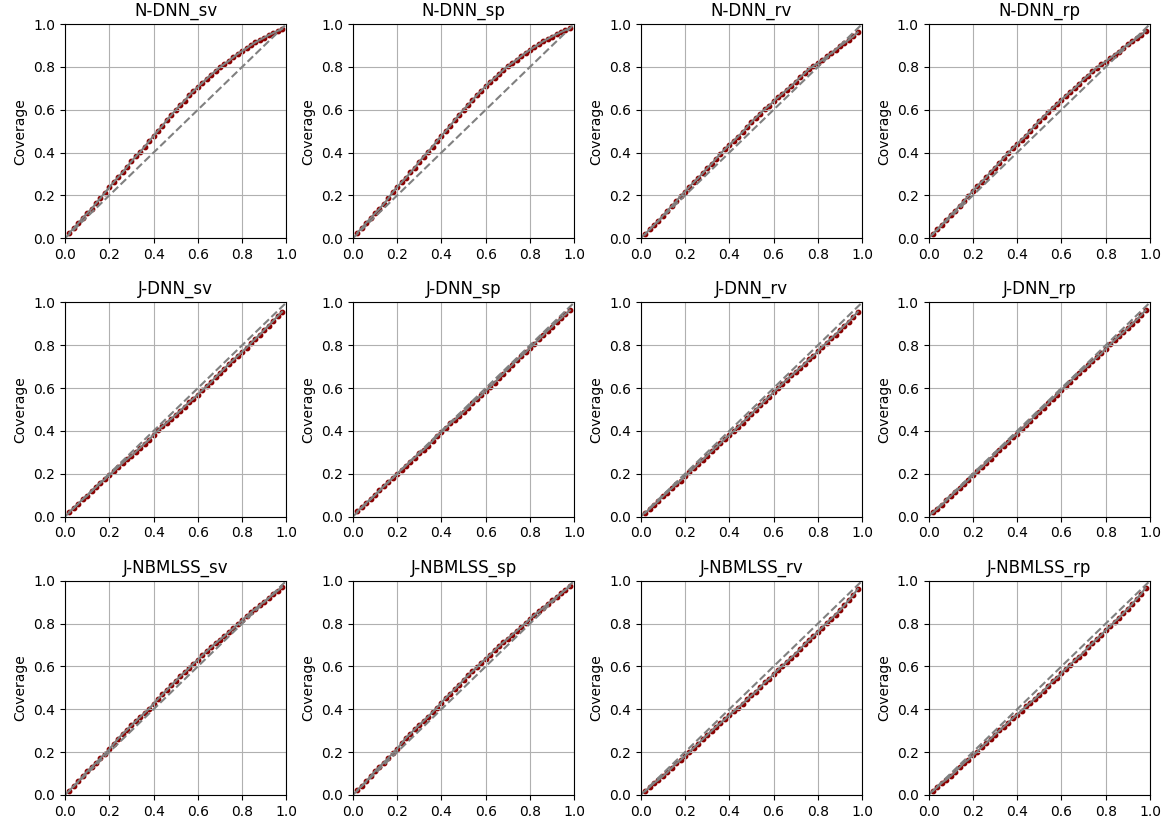}
\end{center}
\caption{Calibration plots - DE TS23 test set}
\label{DE_cali}
\end{figure}

\begin{figure}[t!]
\begin{center}
\includegraphics[width=0.91\linewidth]{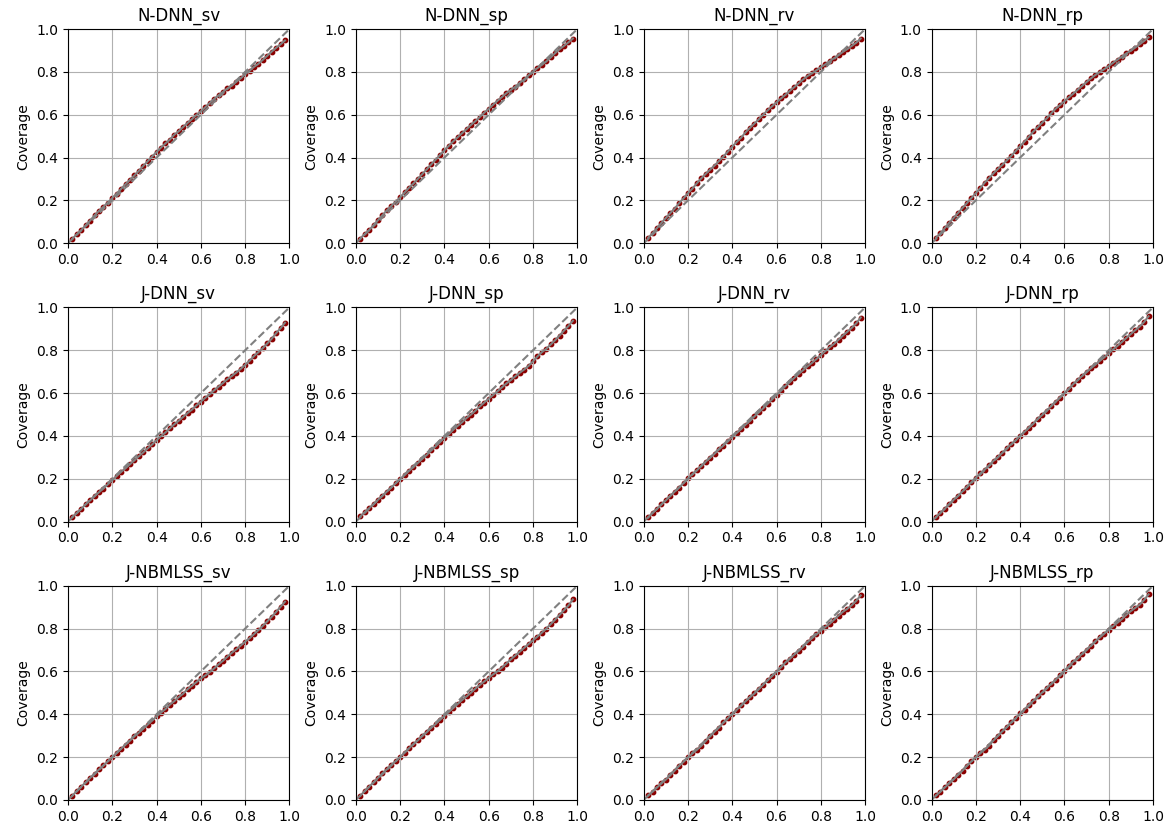}
\end{center}
\caption{Calibration plots - ES TS23 test set}
\label{ES_cali}
\end{figure}
\begin{figure}[t!]
\begin{center}
\includegraphics[width=0.91\linewidth]{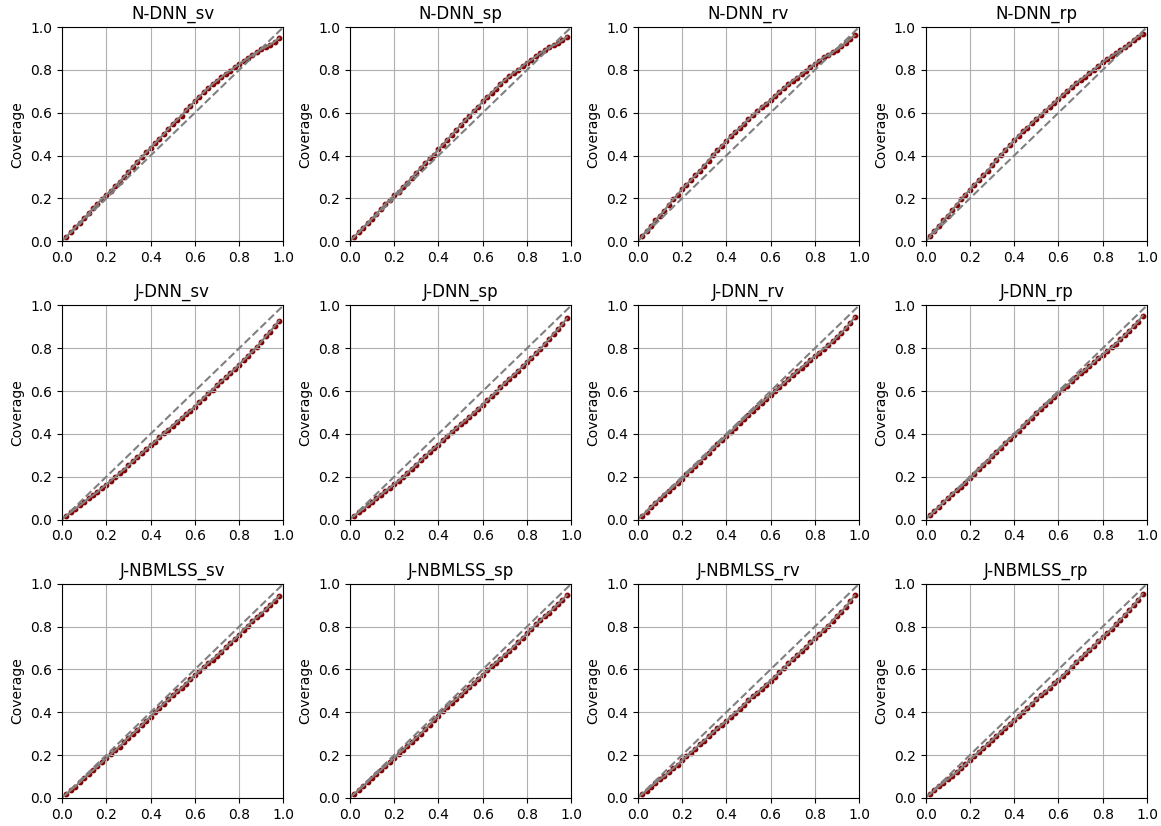}
\end{center}
\caption{Calibration plots - SE TS23 test set}
\label{SE_cali}
\end{figure}

\begin{figure}[t!]
\begin{center}
\includegraphics[width=0.91\linewidth]{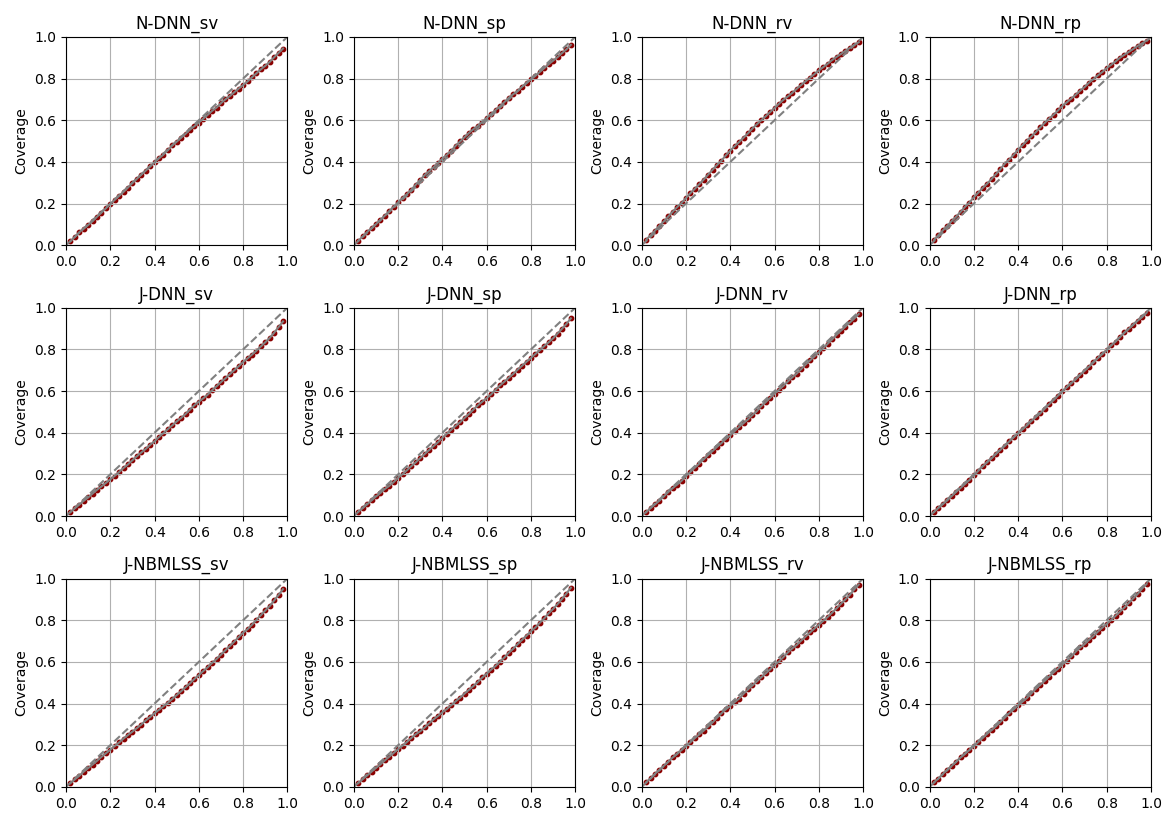}
\end{center}
\caption{Calibration plots - BE TS24 test set}
\label{BE_cali24}
\end{figure}
\begin{figure}[t!]
\begin{center}
\includegraphics[width=0.91\linewidth]{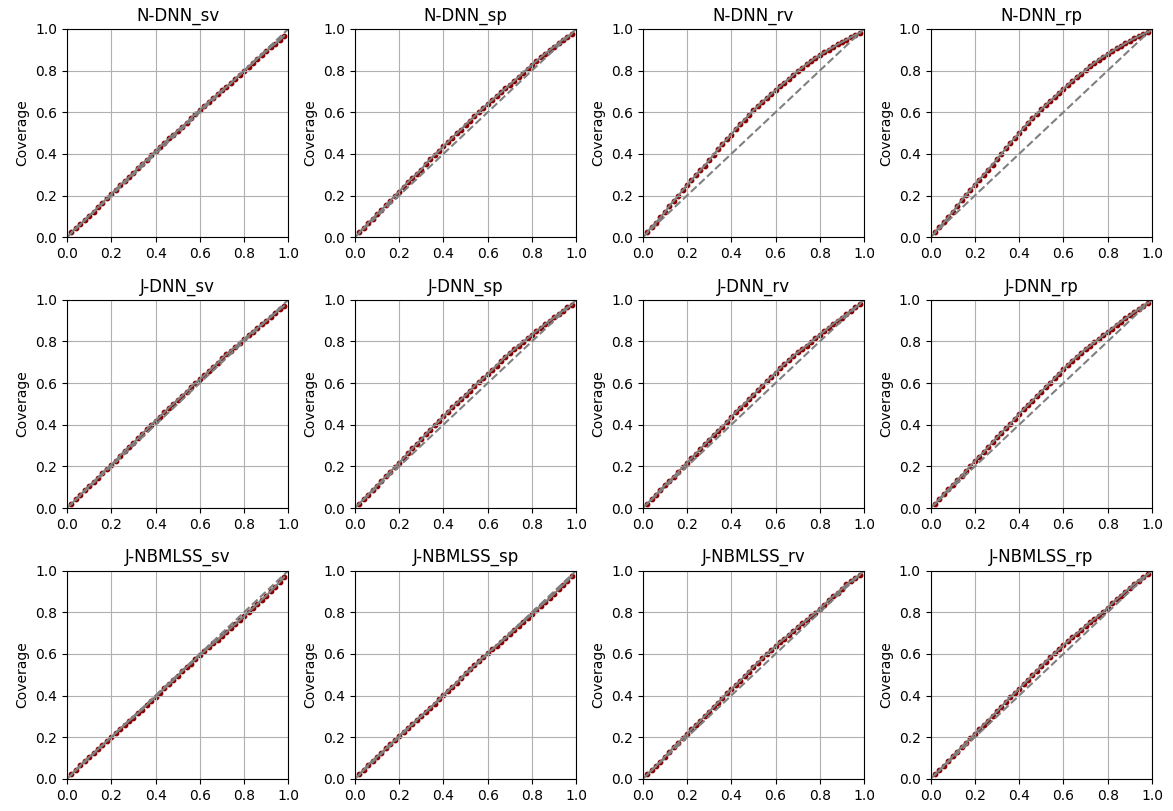}
\end{center}
\caption{Calibration plots - DE TS24 test set}
\label{DE_cali24}
\end{figure}

\begin{figure}[t!]
\begin{center}
\includegraphics[width=0.91\linewidth]{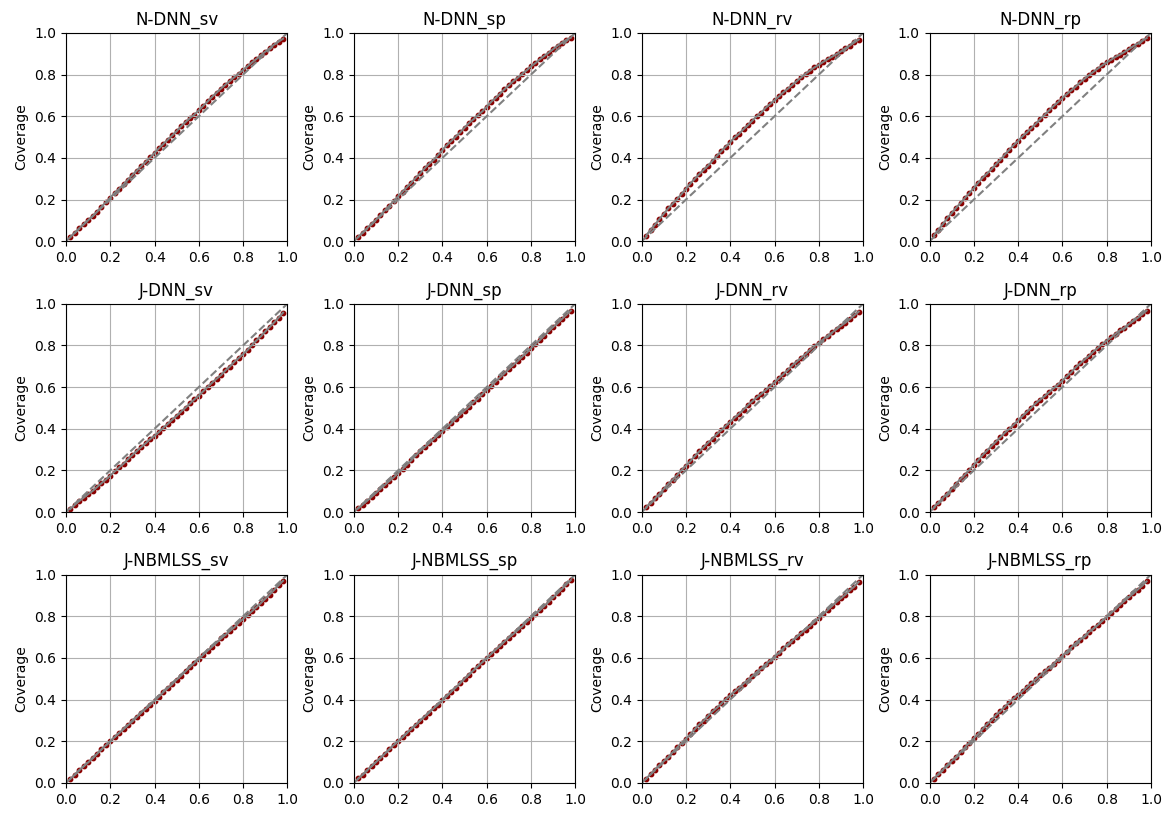}
\end{center}
\caption{Calibration plots - ES TS24 test set}
\label{ES_cali24}
\end{figure}
\begin{figure}[t!]
\begin{center}
\includegraphics[width=0.91\linewidth]{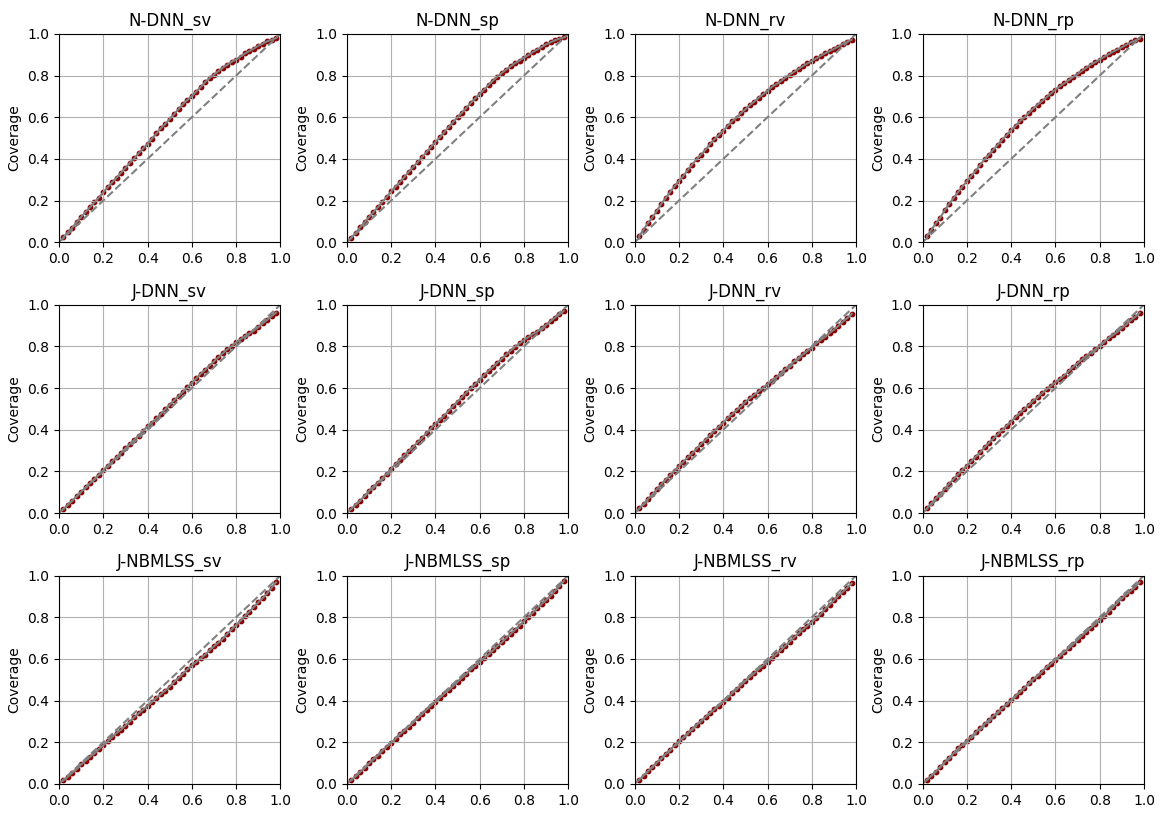}
\end{center}
\caption{Calibration plots - SE TS24 test set}
\label{SE_cali24}
\end{figure}

\begin{figure}[t!]
\begin{center}
\includegraphics[width=0.91\linewidth]{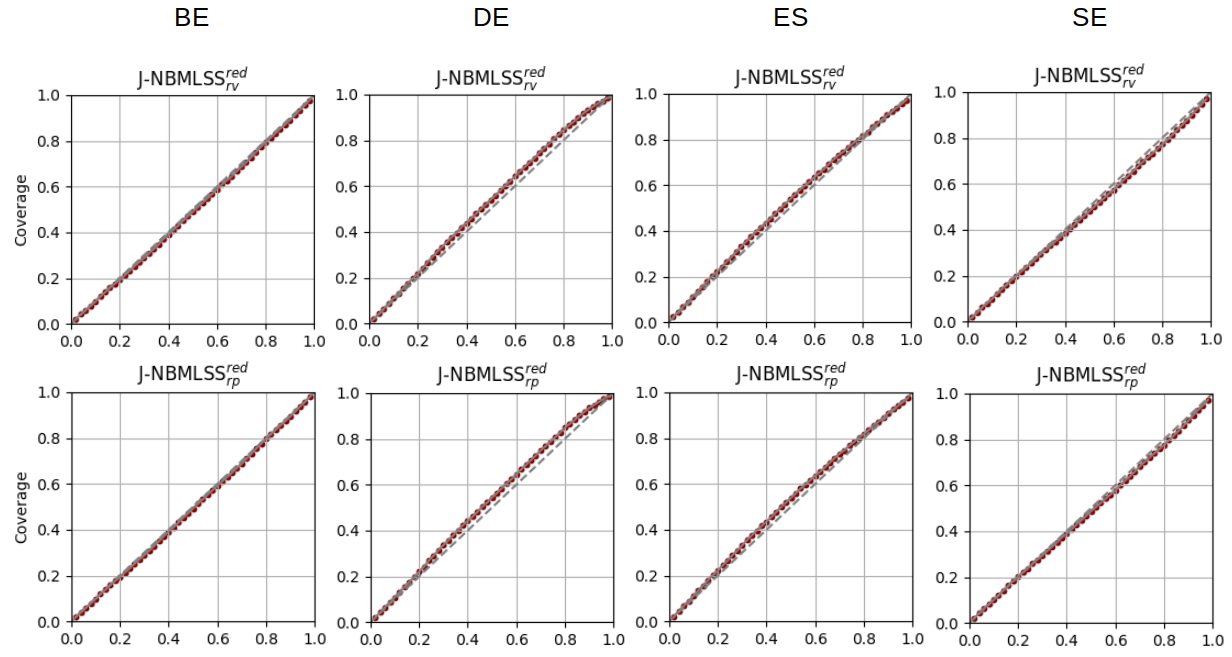}
\end{center}
\caption{Calibration plots - TS24 test set for the masked exogenous configuration}
\label{TS24_cali_masked}
\end{figure}

\begin{figure}[t!]
\begin{center}
\includegraphics[width=0.92\linewidth]{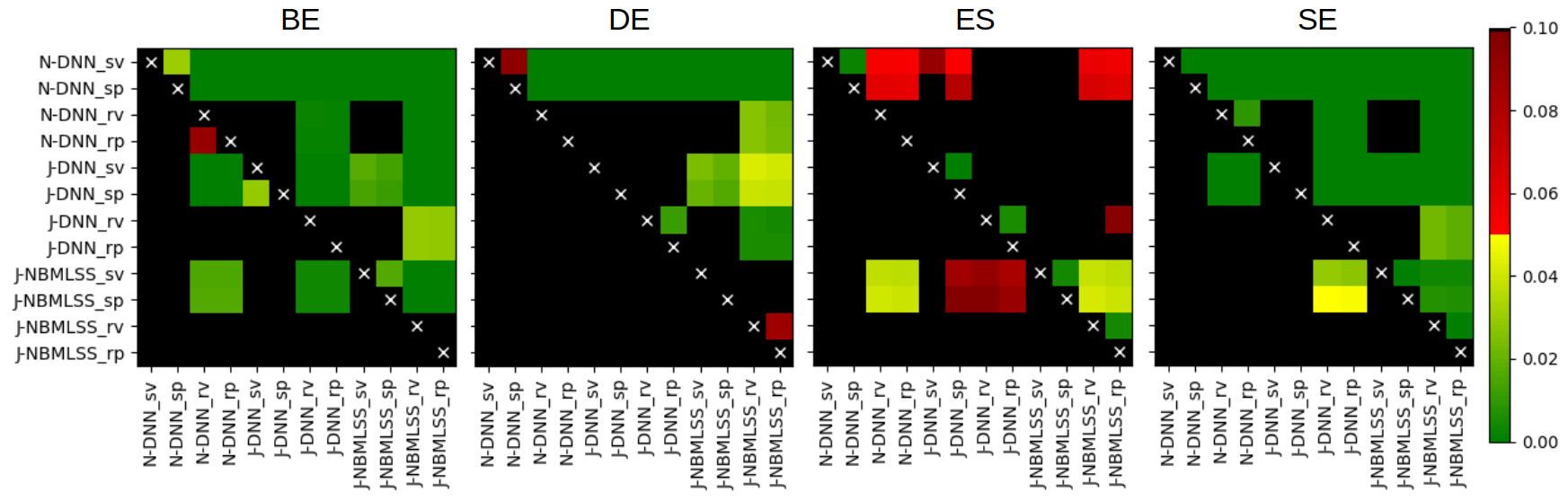}
\end{center}
\caption{DM test on test set TS23 CRPS scores}
\label{DM_Pinball_fig}
\end{figure}

\begin{figure}[t!]
\begin{center}
\includegraphics[width=0.92\linewidth]{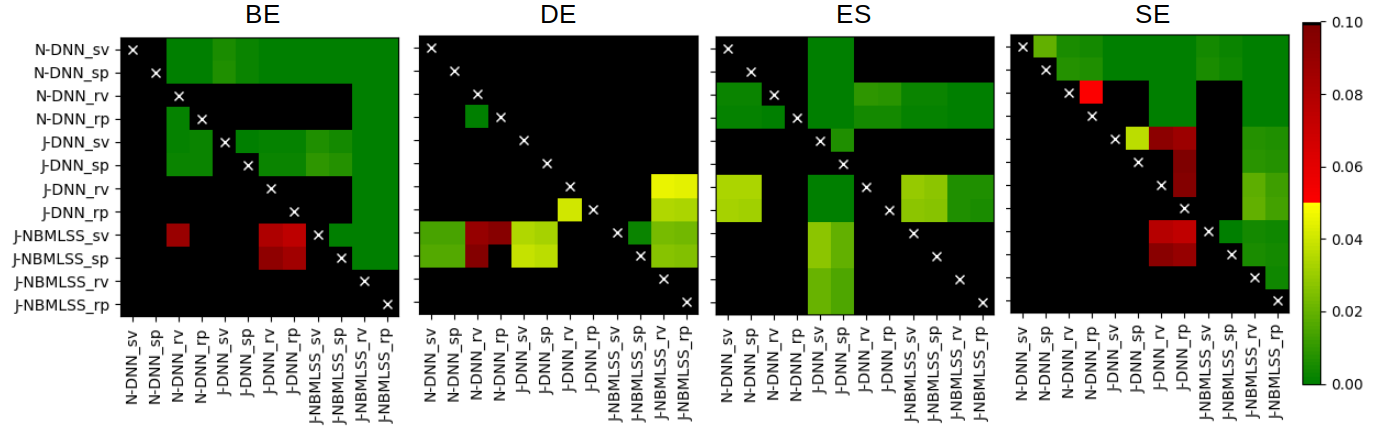}
\end{center}
\caption{DM test on test set TS24 CRPS scores}
\label{DM_Pinball_fig24}
\end{figure}

\begin{figure}[t!]
\begin{center}
\includegraphics[width=0.99\linewidth]{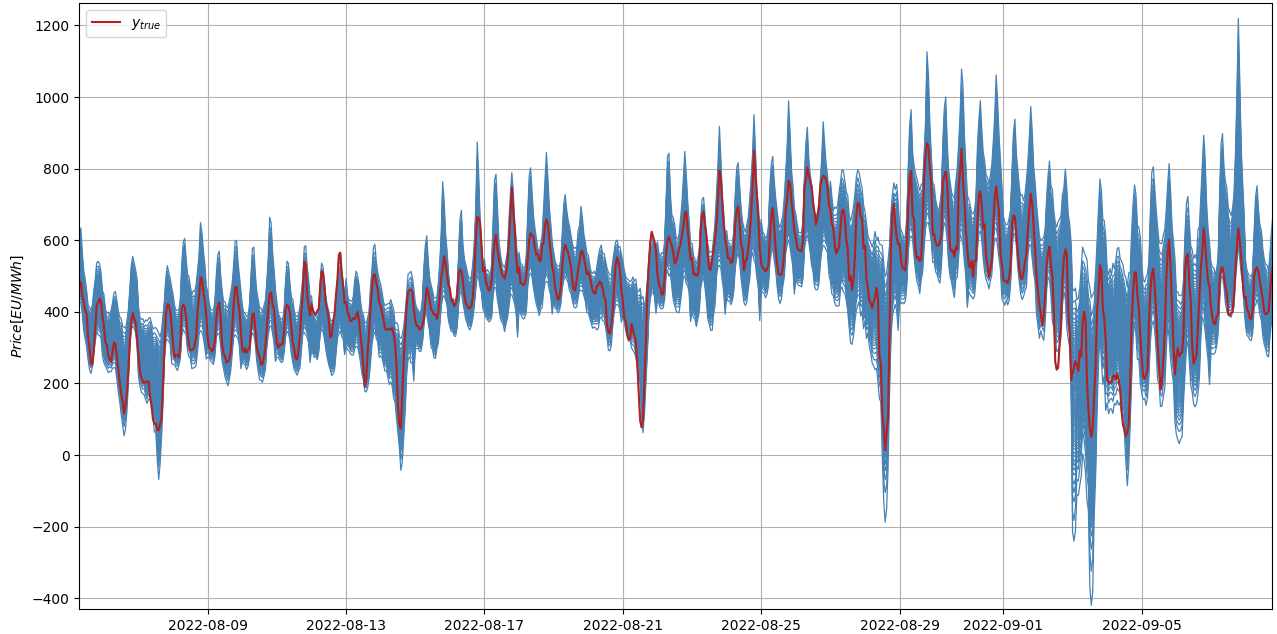}
\includegraphics[width=0.99\linewidth]{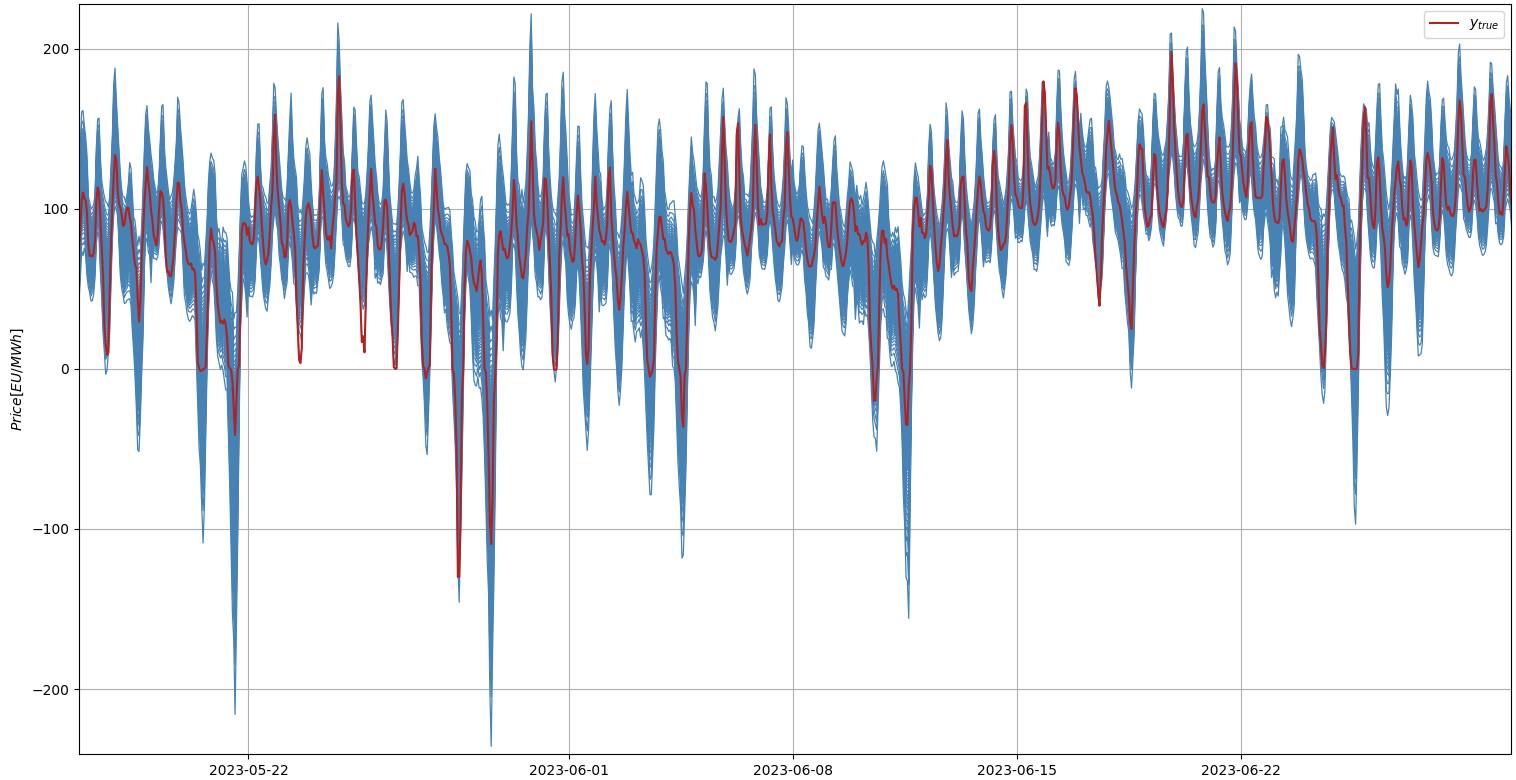}
\end{center}
\caption{Samples of predicted distribution percentiles on DE market TS23}
\label{preds2022-2023}
\end{figure}

\end{document}